\pdfoutput=1

\PassOptionsToPackage{sort}{natbib}

\documentclass[11pt]{article}

\usepackage{acl}
\usepackage{times}
\usepackage{latexsym}
\usepackage[T1]{fontenc}
\usepackage[utf8]{inputenc}
\usepackage{microtype}
\usepackage{inconsolata}

\usepackage{multirow}
\usepackage{multicol}
\usepackage{bbm}
\usepackage{dsfont}
\usepackage[inline]{enumitem} 
\usepackage{amsmath}
\usepackage{mathtools}
\usepackage{url}            
\usepackage{csquotes}
\usepackage{natbib}
\usepackage{graphicx}
\usepackage{subfigure}
\usepackage{acronym}
\usepackage{newfloat}
\usepackage{listings}
\usepackage{algorithm}
\usepackage{algorithmic}
\usepackage{booktabs}
\usepackage{xcolor}

\usepackage[skip=2pt]{caption}

\newcommand{\heading}[1]{\vspace*{.5mm}\noindent\textbf{#1.}}

\title{Pretraining Data Detection for Large Language Models: \\ A Divergence-based Calibration Method}

\begin{document}

\author{Weichao Zhang\textsuperscript{\rm 1,2,3}\space\space 
Ruqing Zhang\textsuperscript{\rm 1,2}\space\space 
Jiafeng Guo\textsuperscript{\rm 1,2}\thanks{Corresponding author}\\ 
\textbf{Maarten de Rijke}\textsuperscript{\rm 4}\space\space 
\textbf{Yixing Fan}\textsuperscript{\rm 1,2}\space\space 
\textbf{Xueqi Cheng}\textsuperscript{\rm 1,2}\space\space\\
\textsuperscript{\rm 1}CAS Key Lab of Network Data Science and Technology, ICT, CAS, Beijing, China\\
\textsuperscript{\rm 2}University of Chinese Academy of Sciences, Beijing, China\\
\textsuperscript{\rm 3}Zhongguancun Laboratory, Beijing, China\\
\textsuperscript{\rm 4}University of Amsterdam, Amsterdam, The Netherlands\\
\{zhangweichao22z, zhangruqing, guojiafeng, fanyixing, cxq\}@ict.ac.cn, m.derijke@uva.nl
}

\maketitle

\begin{abstract}

As the scale of training corpora for large language models (LLMs) grows, model developers become increasingly reluctant to disclose details on their data. 
This lack of transparency poses challenges to scientific evaluation and ethical deployment. 
Recently, pretraining data detection approaches, which infer whether a given text was part of an LLM's training data through black-box access, have been explored. 
The Min-K\% Prob method, which has achieved state-of-the-art results, assumes that a non-training example tends to contain a few outlier words with low token probabilities. 
However, the effectiveness may be limited as it tends to misclassify non-training texts that contain many common words with high probabilities predicted by LLMs. 
To address this issue, we introduce a divergence-based calibration method, inspired by the divergence-from-randomness concept, to calibrate token probabilities for pretraining data detection. 
We compute the cross-entropy (i.e., the divergence) between the token probability distribution and the token frequency distribution to derive a detection score.
We have developed a Chinese-language benchmark, PatentMIA, to assess the performance of detection approaches for LLMs on Chinese text. 
Experimental results on English-language benchmarks and PatentMIA demonstrate that our proposed method significantly outperforms existing methods. Our code and PatentMIA benchmark are available at \url{https://github.com/zhang-wei-chao/DC-PDD}.

\end{abstract}

\section{Introduction}
A critical element contributing to the effectiveness of large language models (LLMs) is the large volume of data used for pretraining.
In many cases, model developers are reluctant to disclose information about their training corpus \cite{GPT4, Llama2, GPT3, Baichuan, Qwen}.
This lack of transparency complicates the assurance that all ethical and legal standards are met.
The pretraining corpus may contain unauthorized private information or copyrighted content \cite{Priuse, cpuse2}.
Indeed, OpenAI and NVIDIA face lawsuits over copyright issues related to their training data \cite{Openaisuit, Nvidiasuit}.
Moreover, a lack of transparency around the pretraining data used prevents us from properly addressing the data contamination problem \cite{CDuse1, CDuse2} and, hence, from determining whether an LLM's performance is due to genuine task understanding or to prior exposure to test data.
We focus on the following key question:
\emph{How can we detect if a black-box LLM was pretrained on a given text, considering that its training data is undisclosed?}

\begin{figure*}[htbp]
  \centering
  \includegraphics[width=1\linewidth]{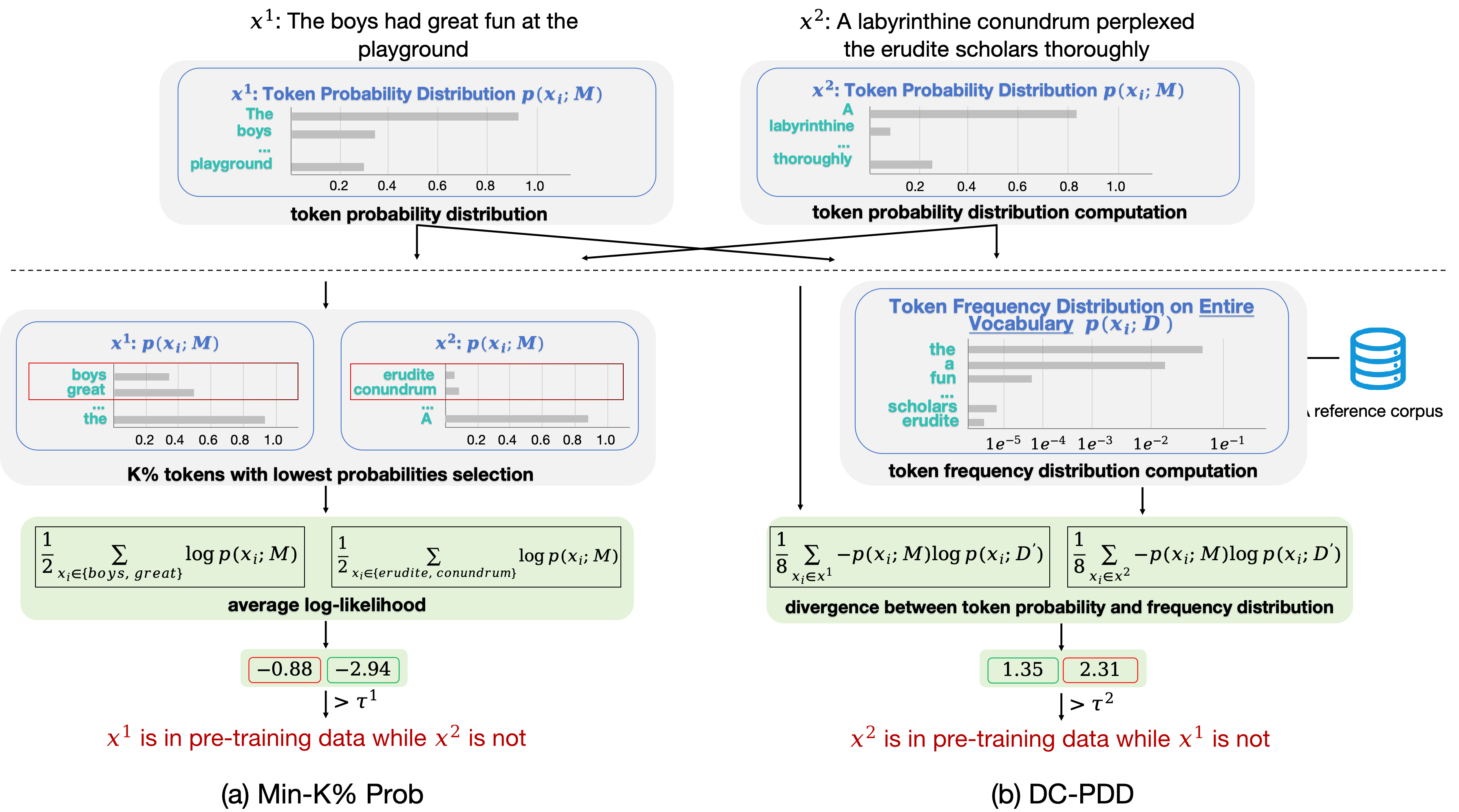}
  \caption{A conceptual example: Let $x^1$ represent a non-training text and $x^2$ a training text. (a) \textbf{Min-K\% Prob} directly selects the $k$\% of tokens with the lowest probabilities for detection. (b)~\textbf{DC-PDD} computes the divergence between the token probability distribution and the token frequency distribution for detection.}
  \label{figure1}
  \vspace{-2mm}
\end{figure*}

The pretraining data detection problem can be viewed as an instance of the membership inference attack (MIA) task \cite{FirstMIA}, where the primary objective is to determine if a particular text was part of a target LLM's training corpus.
Prevailing methods to tackle this problem are  based on the idea that a text's token probability distribution can reveal its inclusion in the training set.
E.g., the Min-K\% Prob method \cite{Min-k} is based on the hypothesis that non-training examples tend to have more tokens assigned lower probabilities than training examples do.
Min-K\% Prob relies on the assumption that data with higher probability is more likely to be training data.
Language models trained with a cross-entropy loss function tend to favor high-frequency tokens when conducting next-token prediction, which will also lead to LLMs generally predicting higher probabilities for high-frequency tokens \cite{CeLoss}.
As the conceptual example shown in figure \ref{figure1}, $x^1$ is a non-training text and $x^2$ is a training text. We can see that the lowest raw token probabilities for $x^1$ are higher than those for $x^2$, which may be because the words in $x^1$ (e.g., "boys", "great") are generally more common than the words in $x^2$ (e.g., "erudite", "conundrum"). Therefore, Min-k\% Prob will calculate a detection score of -0.88 for $x^1$ and -2.94 for $x^2$, which means that $x^1$ is more likely to be considered a training text than $x^2$. This is contrary to the actual situation.

Inspired by the \emph{divergence-from-randomness} theory \cite{DFR}, we introduce a divergence-based calibration method, named DC-PDD, to calibrate the token probabilities for pretraining data detection.
The basic idea underlying divergence-from-randomness is that \emph{the higher the divergence of the within-document term-frequency of a word in a document from its frequency within the collection, the more information the word carries}.
In our scenario, the within-document term-frequency can be interpreted as the target LLM’s predicted probability for each token with regard to the text to be detected, to which we refer as the \emph{token probability distribution}.
The frequency of a word within the collection refers to the frequency of each token in the target LLM’s pretraining corpus, to which we refer as the \emph{token frequency distribution}.
According to the divergence-from-randomness theory, the higher the divergence between these two distributions, the more informative the tokens are in indicating that the text was part of the model's training corpus, rather than solely relying on token probabilities as the indicator for detection.

Like prior works \cite{Min-k, LLMMIAA}, we assume that we only have access to the target LLM as a black box: we can compute token probabilities for the text to be detected but have no access to the internals of the LLM (e.g., weights and activations).
We first obtain the token probability distribution by querying the LLM with the text.
Next, we use a large-scale publicly available corpus as a reference corpus to obtain an estimation of the token frequency distribution since an LLM's pretraining corpus is not accessible usually.
We then calibrate the token probabilities by comparing the token probability distribution to the token frequency distribution.
Based on the calibrated token probabilities, we derive a score for pretraining data detection.  
Finally, a predefined threshold is applied to the score to determine whether the text was included in the LLM’s pretraining corpus.

Figure \ref{figure1}(b) illustrates that DC-PDD assigns a score to text that better reflects whether it is training data or non-training data (i.e., a training text should have a higher score than a non-training text).
In contrast to other calibration methods \cite{Extract, Min-k+}, DC-PDD neither requires additional reference models nor extra access requirements on the target LLM.

\begin{table*}[t]
\centering
\setlength{\tabcolsep}{1.5mm}
\resizebox{\linewidth}{!}{
\begin{tabular}{lll rr l}
\toprule
Benchmark                & Data source                & Language                 & Text length & \#Examples & Applicable models \\ 
\midrule
\multirow{4}{*}{WikiMIA \cite{Min-k}} & \multirow{4}{*}{Wikipedia} & \multirow{4}{*}{English} & 32          & 776        & \multirow{4}{*}{\begin{tabular}[c]{@{}l@{}}Open-source LLMs released between 2017\\    and 2023\end{tabular}}                      \\
                         &                            &                          & 64          & 542        &                                                                                                                         \\
                         &                            &                          & 128         & 250        &                                                                                                                         \\
                         &                            &                          & 256         & 82         &                                                                                                                         \\ \midrule
BookMIA \cite{Min-k}                  & Books                      & English                  & 512         & 9,870       & \begin{tabular}[c]{@{}l@{}}OpenAI models released  before 2023\end{tabular}                                           \\ \midrule
PatentMIA (\textbf{Ours})                & GooglePatent               & Chinese                  & 512         & 10,000      & \begin{tabular}[c]{@{}l@{}}Open-source Chinese LLMs released between \\   January 1, 2023 and March 1, 2024\end{tabular} \\ 
\bottomrule
\end{tabular}
}
\caption{Benchmark summary statistics: Each benchmark has an equal split of training and non-training examples. ``Text Length'' refers to the number of words contained in each text example of the benchmark. ``\#Examples'' denotes the number of text examples in the benchmark.}
\label{table1}
\vspace{-4mm}
\end{table*}

To facilitate this study and the evaluation of pretraining data detection for LLMs, we introduce a new benchmark named PatentMIA, specifically designed for Chinese-language pretraining data detection.
PatentMIA is sourced from Google-Patents \cite{GooglePatents} and constructed following \citet{Min-k}, who distinguish between training and non-training data based on cut-off dates of the target LLM, where training data precedes, and non-training data follows, the cut-off date.

We conduct experiments on two English-language benchmarks \cite{Min-k} and on PatentMIA against a range of representative, state-of-the-art methods.
Our experiments show that the proposed DC-PDD significantly outperforms prior methods.
E.g., in the commonly used detection performance metrics, AUC and TPR@5\%FPR, DC-PDD surpasses Min-K\% Prob by $8.6$\% and $13.3$\%, respectively, on existing BookMIA benchmark.

\section{Problem Statement}

\subsection{Task Description}
Formally, given a piece of text $x$ and an LLM $\mathcal{M}$ with no knowledge of its pretraining corpus $\mathcal{D}$,  the \emph{pretraining data detection task} aims to design a method to determine if $x$ was included in $\mathcal{D}$.
Thus, given $x$ and $\mathcal{M}$ as input, a method $\mathcal{A}$ for the pretraining data detection task returns $1$ if it predicts that $x$ is included in $\mathcal{D}$ and $0$ if it is not:
\begin{equation}
     \mathcal{A}(x, \mathcal{M}) \rightarrow \{0, 1\}.
\end{equation}

\heading{Black-box setting} Like prior works \cite{Min-k, LLMMIAA}, we assume that we have access to $\mathcal{M}$ as a black-box, which means that we can compute token probabilities for $x$. The internals of the model, such as the weights and activations, are not available.

\subsection{Benchmark Construction}
Unlike traditional membership inference attacks  \cite{MLMIA1, FIMIA2, LiRA}, which are conducted on locally trained models where the training and non-training data are explicitly known, the pretraining data detection for LLMs poses a new challenge as the pretraining corpus of LLMs is not disclosed.
Here, we introduce existing benchmarks and our newly constructed benchmark that are specifically designed for LLMs. Table \ref{table1} shows their overall statistics. 

\heading{Pre-existing datasets} \citet{Min-k} proposed a benchmark construction method by distinguishing between the training and non-training data based on the knowledge cut-off date of the target LLM, where training data precedes and non-training data follows the cut-off date. This method has been used to construct two English-language benchmarks: WikiMIA and BookMIA. 
In this paper, we conduct experiments on these benchmarks.

\heading{A Chinese-language benchmark: PatentMIA} Existing benchmarks for the pretraining data detection task are exclusively in English.
Other languages exhibit unique grammatical characteristics such as flexible spacing and case insensitivity compared to English, potentially influencing the effectiveness of methods for the detection task.
These differences warrant specific benchmarks to assess the performance of detection methods in languages other than English. We propose a Chinese-language benchmark for that reason. Next, we detail the construction of the PatentMIA benchmark.

\emph{Data source.} We collect data from Google-Patents \cite{GooglePatents} as (i) it contains a large volume of high-quality, publicly available Chinese patent texts and some publicly available large-scale Chinese corpora like ChineseWebText \cite{CWT} explicitly incorporate data from this website, which indicates that existing LLMs are highly likely to have used such data for pretraining; and (ii) if the priority date of a patent is after the release date of the LLM, there is a guarantee that the patent text was not present during LLM's pretraining. 

\emph{Data collection.} Based on Google-Patents, we construct a Chinese-language benchmark called PatentMIA as follows. 
\begin{enumerate*}[label=(\roman*)]
    \item \textit{Data crawling.} We randomly crawl 5,000 Chinese patent pages with a priority date \textit{after March 1, 2024} and 5,000 patent pages with a publication date \textit{before January 1, 2023} respectively.
    \item \textit{Data preprocessing.} These pages then undergo several preprocessing and cleaning steps similar to those used in ChineseWebText to ensure the data format matches the pretraining data format of LLMs.
    \item \textit{Snippet extraction.} For each page, we randomly extract a snippet of 512 words from the original content, creating a balanced set of 10,000 examples. We use \textit{jieba}\footnote{\url{https://github.com/fxsjy/jieba}} to segment Chinese texts into words.
\end{enumerate*}

\section{Method}

\vspace{-1mm}
\subsection{Overview}
Given a piece of text $x=x_1x_2\ldots x_n$, where $x_i$ represent the tokens after tokenizing $x$, and a target LLM $\mathcal{M}$, we compute a detection score by measuring the divergence between the token probability distribution of $x$ and the token frequency distribution in pretraining corpus, without any model training processes. Our method consists of four steps:
\begin{enumerate*}[label=(\roman*)]
    \item \textit{Token probability distribution computation}, by querying $\mathcal{M}$ with $x$ (Section~\ref{subsection:3.2}).
    \item \textit{Token frequency distribution computation}, by using a large-scale publicly available corpus $\mathcal{D'}$ as a reference corpus to obtain an estimation of the token frequency distribution since $\mathcal{M}$'s pretraining corpus is not assumed to be accessible (Section~\ref{subsection:3.3}).
    \item \textit{Score calculation via comparison}, by comparing the above two distributions to calibrate the token probability for each token $x_i$ in $x$, and derive a score for pretraining data detection based on the calibrated token probabilities (Section~\ref{subsection:3.4}).
    \item \textit{binary decision}, by applying a predefined threshold to the score, we predict whether $x$ was included in $\mathcal{M}$’s pretraining corpus or not (Section~\ref{subsection:3.5}).
\end{enumerate*}

We summarize our method in Algorithm~\ref{alogrithm}. 

\begin{algorithm}[!t]
    \caption{Our DC-PDD}\label{alogrithm}
    \renewcommand{\algorithmicrequire}{\textbf{Input:}}
    \renewcommand{\algorithmicensure}{\textbf{Output:}}
    \begin{algorithmic}[1]        
        \REQUIRE A text to be detected $x=x_1x_2\ldots x_n$, a target LLM $\mathcal{M}$, vocabulary of LLM $V=\{x_i \}_{i=1}^{\lvert V \rvert}$, reference corpus $\mathcal{D'}$,  decision threshold $\tau$
        
        \STATE Prepend a start-of-sentence token to $x$
        \FOR{$i=1$ to $n$}
            \STATE Access the token probability $p(x_i; \mathcal{M})$ from $\mathcal{M}$, w.r.t. Eq.~\eqref{eq:3}
        \ENDFOR

        \FOR{$i=1$ to $\lvert V \rvert$}
            \STATE Compute the token frequency $p(x_i; \mathcal{D'})$ based on $\mathcal{D'}$, w.r.t. Eq.~\eqref{eq:5}
        \ENDFOR
        \FOR{$i=1$ to $n$}
            \STATE Compute $\alpha_i$ for $x_i$ based on $p(x_i; \mathcal{M})$ and $p(x_i; \mathcal{D'})$, w.r.t. Eq.~\eqref{eq:6}, \eqref{eq:7}
        \ENDFOR

        \STATE Select $\alpha_i$ corresponding to tokens with the first occurrence in $x$ to compute a score $\beta$, w.r.t. Eq.~\eqref{eq:8}
        \IF{$\beta \geq \tau$}
            \STATE 1: $\mathcal{M}$ was pretrained on $x$
        \ELSE
            \STATE 0: $\mathcal{M}$ was not pretrained on $x$
        \ENDIF
    \end{algorithmic}
    
\end{algorithm}

\subsection{Token Probability Distribution Computation}
\label{subsection:3.2}

To obtain all the probabilities of $x_i$ in $x$ from $\mathcal{M}$, we first prepend a start-of-sentence token, denoted as $x_0$, to $x$, since the model does not return a prediction for the first token:
\begin{equation}
     x'=x_0x_1x_2\ldots x_n.
     \label{eq:2}
\end{equation}
Subsequently, we feed $x'$ into $\mathcal{M}$, resulting in a sequence of predicted probabilities corresponding to the true tokens:
\begin{equation}
     \{p(x_i \mid x_{<i}; \mathcal{M}) : 0 < i \leq n\}.
     \label{eq:3}
\end{equation}
Note that the probability of each token $x_i$ is predicted by $\mathcal{M}$ based on the preceding context $ x_{<i}$ for $0<i \leq n$. For brevity in subsequent expressions, we simplify $p(x_i \mid x_{<i}; \mathcal{M})$ to $p(x_i; \mathcal{M})$.

\subsection{Token Frequency Distribution Computation}
\label{subsection:3.3}

According to the divergence-from-randomness theory, after obtaining the token probability distribution for $x$ from $\mathcal{M}$, we also need to calculate the frequency of $x_i$ appearing in the pretraining corpus $\mathcal{D}$ of $\mathcal{M}$ to get the token frequency distribution. 
However, since $\mathcal{D}$ is not accessible, we cannot directly calculate these terms. 
To address this, we use a large-scale publicly available corpus $\mathcal{D'}$ to obtain an estimation of these terms:
\begin{equation}
     p(x_i; \mathcal{D'})=\frac{\operatorname{count}(x_i)}{N'},
     \label{eq:4}
\end{equation}
where $\operatorname{count}(x_i)$ is the number of occurrences of $x_i$ in $\mathcal{D'}$, and $N'$ is the total number of tokens in $\mathcal{D'}$. 
We employ Laplace smoothing to address the zero probability problem when $x_i$ does not occur in $\mathcal{D'}$ even once:
\begin{equation}
     p(x_i; \mathcal{D'})=\frac{\operatorname{count}(x_i)+1}{N' + \lvert V \rvert },
     \label{eq:5}
\end{equation}
where $\lvert V \rvert$ represents the vocabulary size of  $\mathcal{M}$, i.e., the number of categories of tokens.

\subsection{Score Calculation through Comparison}
\label{subsection:3.4}

We compute the cross-entropy (i.e., the divergence) between the token probability distribution $p(x_i; \mathcal{M})$ and the token frequency distribution $p(x_i; \mathcal{D}')$ to obtain a score $\alpha_i$ for each token $x_i$:
\begin{equation}
     \alpha_i = - p(x_i; \mathcal{M}) \cdot \log p(x_i; \mathcal{D}').
     \label{eq:6}
\end{equation}
We set a hyperparameter $a$ to control the upper bound of $\alpha_i$, preventing the final score from being dominated by a few tokens:
\begin{equation} 
     \alpha_i = 
     \begin{cases}
        \alpha_i, & \text{if }\alpha_i < a \\
        a, & \text{if }\alpha_i \geq a.
    \end{cases}
    \label{eq:7}
\end{equation}
Typically, for a word that appears multiple times in a text, LLMs predict a higher probability for that word in subsequent occurrences since the model has seen the word earlier in the text.
Therefore, we adopt a simple countermeasure that only uses $\alpha_i$ corresponding to the first occurrence of $x_i$ in $x$ to calculate the final score $\beta$:
\begin{equation} \label{eq:8}
     \beta = \frac{1}{\lvert \operatorname{FOS}(x) \rvert} \sum_{x_j \in \operatorname{FOS}(x)} {\alpha_j},
\end{equation}
where $\operatorname{FOS}(x)$ denotes the set of tokens with the first occurrence in $x$.

\subsection{Binary Decision}
\label{subsection:3.5}
After calculating the score $\beta$ for $x$ following the aforementioned three steps, we predict whether $x$ was included in $\mathcal{M}$'s pretraining corpus $\mathcal{D}$ by applying a predefined threshold $\tau$ to $\beta$:
\begin{equation}
    \operatorname{Decision}(x, \mathcal{M})=
    \begin{cases}
        0~(x \notin \mathcal{D}),& \text{if }\beta < \tau \\
        1~(x \in \mathcal{D}),& \text{if }\beta \geq \tau.
    \end{cases}
    \label{eq:9}
\end{equation}
If $\beta$ is not less than $\tau$, we predict that $x$ was included in $\mathcal{D}$; otherwise, it was not.

\section{Experimental Settings}

\textbf{Benchmarks and models.} 
To evaluate the performance of DC-PDD, we conduct experiments on three benchmarks mentioned in Table \ref{table1}.
Specifically, for WikiMIA, we consider OPT-6.7B \cite{OPT}, Pythia-6.9B \cite{Pythia}, Llama-13B \cite{Llama}, and GPT-NeoX-20B \cite{GPT-neox}, since they were released after 2017 and before 2023, and are well-known for incorporating Wikipedia dumps into their pretraining data. For BookMIA, we consider GPT-3,\footnote{davinci-002, an OpenAI model released before 2023, also belongs to the applicable models for BookMIA; text-davinci-003 was used by \citet{Min-k} but it has been deprecated by OpenAI.} since it's an OpenAI model released before 2023. These settings are akin to \citet{Min-k}. For our benchmark PatentMIA, we select Baichuan-13B \cite{Baichuan} and Qwen1.5-14B \cite{Qwen1.5}, since they are representative models in Chinese text generation and are released between January 1, 2023 and March 1, 2024.

\heading{Baselines}
We consider the following methods as our baselines, each predicting whether an example was included in training set based on:
\begin{enumerate*}[label=(\roman*)]
    \item \emph{PPL}: The perplexity of the example.
    \item \emph{Lowercase}: The ratio of the example's perplexity to that of the lowercased example.
    \item \emph{Zlib}: The ratio of the example’s perplexity against its zlib entropy.
    \item \emph{Small Ref}: The ratio of an example's perplexity to the example's perplexity under a smaller model pretrained on the same data.
    \item \emph{Min-K\% Prob} \cite{Min-k}: The average log-likelihood of the $k$\% of tokens with the lowest probabilities. 
    \item \emph{Min-K\%++ Prob} \cite{Min-k+}: The average normalized log-likelihood of the k\% of tokens with the lowest normalized probabilities, where the normalization is based on the statistics of the categorical distribution over the entire vocabulary.
\end{enumerate*}
Note that the first four baselines were introduced in \cite{Extract}. For more details on our baselines, please refer to Appendix~\ref{appendix-baselines}.

\heading{Evaluation metrics} 
Following most existing works \cite{Min-k, LLMMIAA, Min-k+}, we use AUC score (area under ROC curve) and TPR (true positive rate) at a low FPR (false positive rate) (TPR@5\%FPR) as our metrics.
For more details on these metrics, please refer to Appendix~\ref{appendix-metrics}.

\begin{table*}[t]
\centering
\small 
\setlength{\tabcolsep}{4.pt} 
\renewcommand{\arraystretch}{1.1} 
\begin{tabular}{l c cc cccc}
\toprule
\multirow{2}{*}{Method} & BookMIA\phantom{X}        & \multicolumn{2}{c}{PatentMIA} & \multicolumn{4}{c}{WikiMIA}                
\\ 
\cmidrule(r){2-2} 
\cmidrule(r){3-4} 
\cmidrule{5-8} 
 & GPT-3           & Baichuan-13B   & Qwen1.5-14B\phantom{x}   & OPT-6.7B       & Pythia-6.9B    & Llama-13B      & GPT-NeoX-20B   
\\ 
\midrule
PPL                     & 0.635          & 0.608          & 0.599          & 0.625          & 0.651          & 0.678          & 0.707          \\
Lowercase               & 0.671          & -              & -              & 0.587          & 0.605          & 0.606          & 0.680          \\
Zlib                    & 0.537          & 0.634          & 0.618          & 0.644          & 0.676          & 0.697          & 0.723          \\
Small Ref               & -              & 0.657          & 0.565          & 0.654          & 0.660          & 0.658          & 0.714          \\
Min-K\% Prob            & 0.639          & 0.643          & 0.637          & 0.674          & 0.695          & 0.715          & 0.756          \\
Min-K\%++ Prob          & -              & 0.625          & 0.630          & \textbf{0.692}          & 0.697          & \textbf{0.838} & 0.754              \\
DC-PDD                  & \textbf{0.725*} & \textbf{0.699*} & \textbf{0.689*} & 0.677 & \textbf{0.698} & 0.697          & \textbf{0.766*} 
\\
\bottomrule
\end{tabular}
\caption{AUC scores for detecting pretraining texts. \textbf{Bold} indicates the best performing method. Two-tailed t-tests show that DC-PDD significantly improves over Min-K\% Prob ( * indicates $p \leq 0.05$).}
\label{table2}
\end{table*}

\begin{table*}[t]
\centering
\small 
\setlength{\tabcolsep}{4pt} 
\renewcommand{\arraystretch}{1.1} 
\begin{tabular}{l c cc cccc}
\toprule
\multirow{2}{*}{Method} & BookMIA\phantom{X}        & \multicolumn{2}{c}{PatentMIA} & \multicolumn{4}{c}{WikiMIA} 
\\
\cmidrule(r){2-2} 
\cmidrule(r){3-4} 
\cmidrule{5-8} 
& GPT-3           & Baichuan-13B   & Qwen1.5-14B\phantom{x}   & OPT-6.7B       & Pythia-6.9B    & Llama-13B      & GPT-NeoX-20B   
\\ 
\midrule
PPL                     & 0.224          & 0.166             & 0.159             & 0.130          & 0.144          & 0.216          & 0.180          \\
Lowercase               & 0.240          & -                 & -                 & 0.094          & 0.130          & 0.158          & 0.130          \\
Zlib                    & 0.192          & 0.159             & 0.129             & 0.180          & 0.209          & 0.187          & 0.223          \\
Small Ref               & -              & 0.211             & 0.078             & 0.122          & 0.158          & 0.151          & 0.187          \\
Min-K\% Prob            & 0.203          & 0.170             & 0.163             & 0.166          & 0.180          & 0.201          & 0.216          \\
Min-K\%++ Prob          & -              & 0.130             & 0.141             & \textbf{0.215} & 0.201          & \textbf{0.381} & 0.245          \\
DC-PDD                  & \textbf{0.336*} & \textbf{0.264*}  & \textbf{0.271*}      & 0.180*          & \textbf{0.245*} & 0.230*          & \textbf{0.317*} 
\\
\bottomrule
\end{tabular}
\caption{TPR@5\%FPR scores for detecting pretraining texts. \textbf{Bold} indicates the best performing method. Two-tailed t-tests show that DC-PDD significantly improves over Min-K\% Prob ( * indicates $p \leq 0.05$).}
\label{table3}
\vspace{-2mm}
\end{table*}

\heading{Implementation details} 
For the start-of-sentence token $x_0$ to prepend, we use \textit{<|endoftext|>} in Pythia, Qwen1.5, GPT-NeoX and GPT-3, \textit{<s>} in OPT and Llama, and \textit{</s>} in Baichuan.
For the reference corpus $\mathcal{D'}$ to compute the token frequency distribution, we take a subset of C4 \cite{C4} ($\approx 15$Gb) for English text detection and take a subset of ChineseWebText \cite{CWT} ($\approx 15$Gb) for Chinese text detection.
For hyperparameter $a$ settings, we set it to $0.01$ for WikiMIA and PatentMIA detection tasks, and to $10$ for BookMIA.
Since we take the AUC score as our evaluation metric, we do not need to determine a specific threshold $\tau$ in our method.
For the baseline implementation, we set $k = 20$ to achieve the optimal performance of Min-K\% Prob following \citet{Min-k}.
Correspondingly, the hyperparameter $k$ in Min-K\%++ Prob is also set to $20$ for fair comparison.
For the smaller reference model setting, we employ OPT-350M as the smaller model for OPT-6.7B, Pythia-70M for Pythia-6.9B, Llama-7B for Llama-13B, GPT-Neo-125M for GPT-NeoX-20B, Baichuan-7B for Baichuan-13B and Qwen1.5-7B for Qwen1.5-14B.

\section{Experimental Results}

Here, we report our main results, several ablation studies, and additional experiments investigating factors influencing detection performance.

\subsection{Main Results}
Our results can be found in Table \ref{table2} and \ref{table3}. 
We observe that: 
(i) DC-PDD surpasses most baselines across three benchmarks and various target models. For instance, on existing BookMIA benchmark, DC-PDD exceeds the best baseline Lowercase $5.4$\% and $9.6$\% in terms of AUC and TPR@5\%FPR. On our PatentMIA benchmark, DC-PDD exceeds the best baseline Min-K\% Prob $5.4$\% and $13.2$\% in terms of AUC and TPR@5\%FPR.
(ii) Compared to Min-K\% Prob, the AUC improvement of DC-PDD on the WikiMIA benchmark is less than that of Min-K\%++ Prob, possibly because WikiMIA has only 250 examples, with fewer cases shown in Figure \ref{figure1} we aim to optimize. While Min-K\%++ Prob calibrates token probabilities from other points, which might suit these examples better. This indicates that token probabilities are impacted by various factors and are unreliable for detection. Hence, we plan to explore better detection signals in the future.
(iii) The superior performance of DC-PDD is more agnostic to data and models, in comparison to other methods. For example, while Min-K\% Prob and Min-K\%++ Prob perform well on models using the WikiMIA benchmark, they do not do as well on models using the PatentMIA benchmark. A similar phenomenon can be observed with the Zlib method.
(iv) Additionally, the Small Ref method are not applicable to GPT-3, as closed-source models lack corresponding smaller models in the same series. The Min-K\%++ Prob is also not applicable to GPT-3 since GPT-3 do not provide the access to the next-token prediction probability distribution across the model’s entire vocabulary. The Lowercase method is unsuitable for detecting Chinese text, as Chinese characters do not have case distinctions. 
(v) By evaluating performance on the PatentMIA benchmark, except for the Lowercase method, it is evident that existing methods are still effective for Chinese-language pretraining data detection, with our method consistently achieving the best results.

\begin{figure}[t]
  \centering
  \includegraphics[width=1\linewidth]{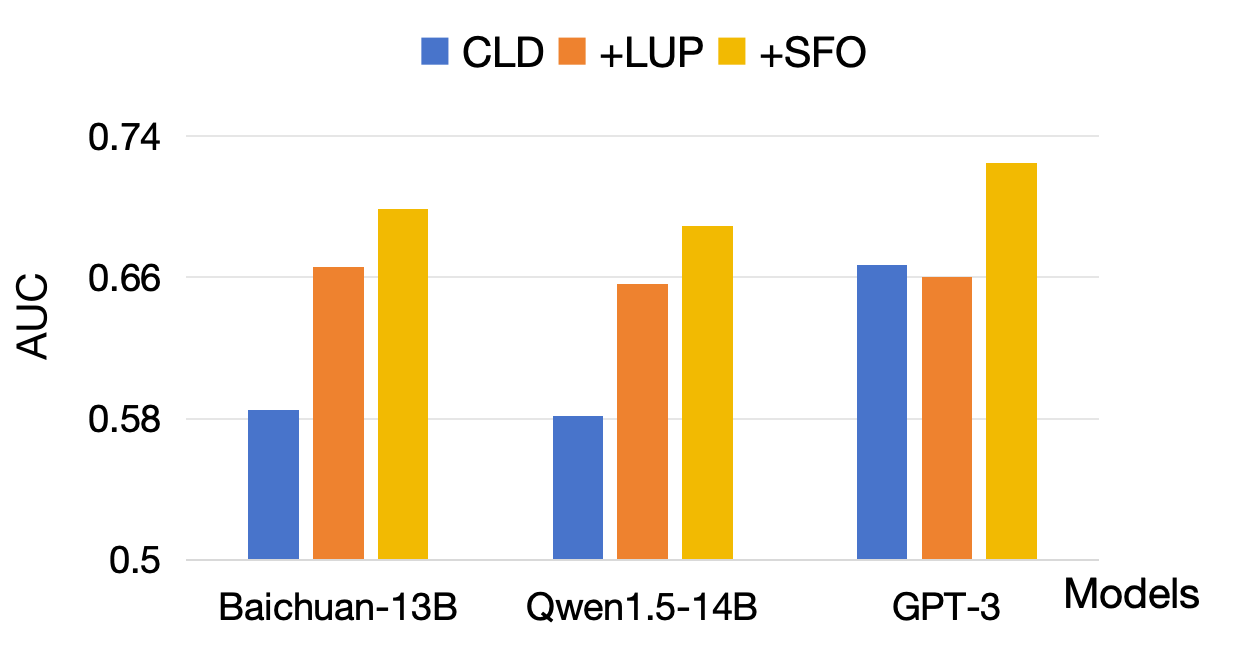}
  \caption{Ablation studies of DC-PDD}
  \label{figure3}
  \vspace{-4mm}
\end{figure}

\vspace{-2mm}
\subsection{Ablation Studies}
DC-PDD employs two strategies before using the calibrated token probabilities to compute the score $\beta$ for $x$ for detection. They are 
(i) \textbf{LUP}: \textbf{L}imiting the \textbf{UP}per bound of each calibrated token probability, w.r.t. Eq.~\eqref{eq:7}, and
(ii) \textbf{SFO}: only \textbf{S}electing the calibrated token probabilities corresponding to tokens with the \textbf{F}irst \textbf{O}ccurrence in $x$ to compute $\beta$, w.r.t. Eq.~\eqref{eq:8}.
We conduct ablation studies to explore the effect of these strategies using the following three method variants:
\begin{itemize}[leftmargin=*,nosep]
    \item \textbf{CLD}: It serves as the initialization of DC-PDD by averaging all the \textbf{C}a\textbf{L}ibrate\textbf{D} token probabilities to compute a score for detection.
    \item +\textbf{LUP}: Based on ‘CLD’, it incorporates the LUP strategy to compute $\beta$.
    \item +\textbf{SFO}: Based on ‘+LUP’, it further incorporates the SFO strategy to compute $\beta$.
\end{itemize}

Results are shown in Figure \ref{figure3}.
For Baichuan-13B and Qwen1.5-14B, both strategies contribute to the effectiveness of DC-PDD.
However, for GPT-3, we found that the LUP strategy did not result in a significant performance improvement.
We speculate that this may be related to the setting of the hyperparameter $a$ involved in the LUP strategy.
Therefore, we discuss the impact of $a$ on DC-PDD in detail in Section~\ref{section5.3}.

\begin{figure}[t]
	\centering
	\subfigure[AUC score vs. model size.] {\includegraphics[width=.48\textwidth]{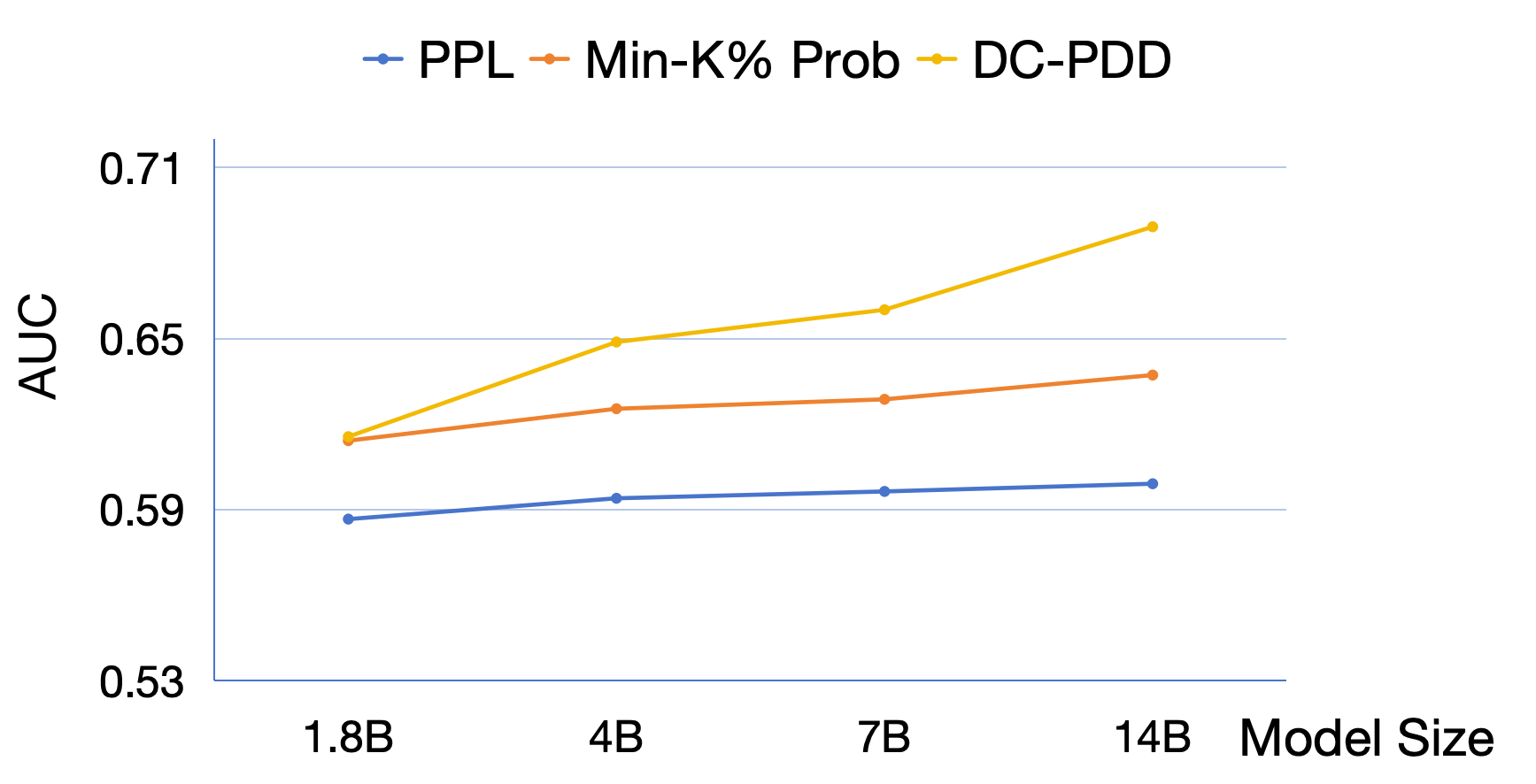}}
	\subfigure[AUC score vs. text length.] {\includegraphics[width=.48\textwidth]{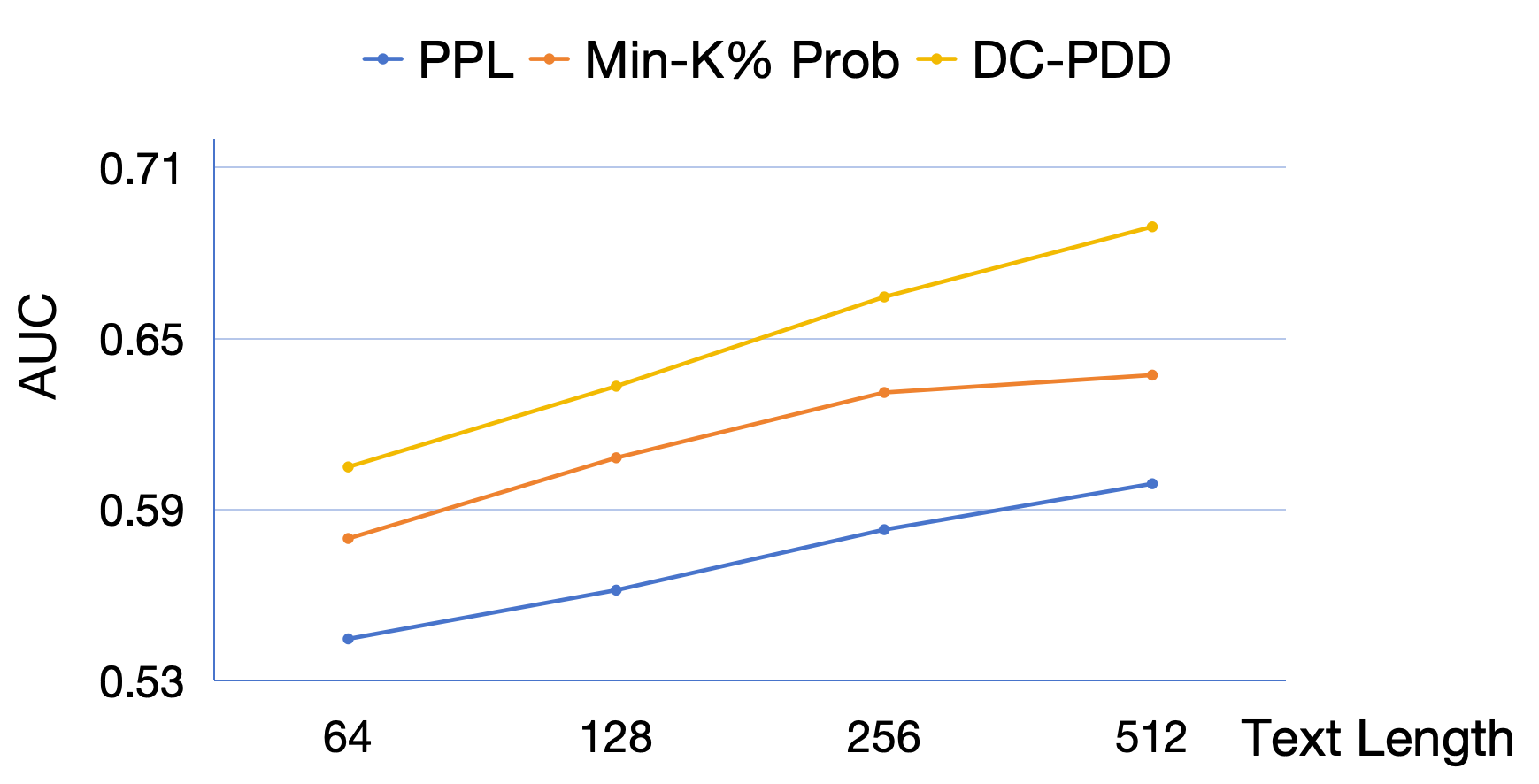}}
	\caption{The performance of DC-PDD w.r.t model size and text length.}
	\label{figure2}
    \vspace{-3mm}
\end{figure}

\subsection{Impact of Different Factors} \label{section5.3}
This section explores several factors that may influence the performance of DC-PDD, including two method-independent factors (model size and text length) and two method-dependent factors (the reference corpus $\mathcal{D'}$ and hyperparameter $a$).

\heading{Model size} To investigate the impact of model size on the performance of DC-PDD, we analyze the Qwen1.5 family with models of 1.8B, 4B, 7B, and 14B versions to determine if larger models demonstrate improved results. As illustrated in Figure \ref{figure2}(a), DC-PDD consistently achieves the best results across all model sizes, and like other methods, the AUC score increases as the model size grows, confirming findings from prior research \cite{Min-k, wb}. The reason for this trend is probably because larger models, having more parameters, are better at memorizing the pre-training data.

\heading{Text length} We further explore the potential impact of text length on the performance of DC-PDD. For this purpose, we perform assessments using four different length settings (64, 128, 256, 512) in our PatentMIA benchmark to determine whether short texts are more challenging than longer texts. Figure \ref{figure2}(b) illustrates that DC-PDD still consistently outperforms other baselines across all text length settings, and the AUC score also improves with increasing length in Chinese-language pretraining data detection. This trend may be due to the fact that longer texts carry more information that the target model has memorized, making them easier to differentiate from non-training texts. 

\heading{Reference corpus $\mathcal{D'}$} Recall that we use a reference corpus $\mathcal{D'}$ to estimate the token frequency distribution of the LLM's pretraining corpus, w.r.t. Eq.~\eqref{eq:4}.
To analyze the effect of different reference corpora on the efficacy of the method, we compare the performance of DC-PDD under various reference corpus settings across different scales and domains.
Specifically, when detecting WikiMIA-128 from pythia-6.9B, we employee $\approx 1$Gb of C4 corpus, $\approx 10$Gb of C4 corpus, $\approx 1$Gb of Case-law corpus, and $\approx 10$Gb of Case-law corpus as the reference corpus respectively. Note that the Case-law \cite{case-law} is a corpus in the legal domain.
As shown in Table \ref{table4}, We observe that the performance of DC-PDD does not exhibit significant differences across the various reference corpora, indicating that DC-PDD is not sensitive to the selection of a reference corpus. 
Notably, when the reference corpus is chosen as the $\approx 10$Gb of C4 corpus, the performance of DC-PDD is the best. This enhancement may be attributed to the greater diversity of the C4 corpus compared to the $\approx 10$Gb of Case-law corpus, as well as the richer data compared to the $\approx 1$Gb of C4 corpus, which allow for a more accurate estimation of the token frequency distribution in the LLM's pretraining corpus, thereby resulting in better performance.

\begin{table}[t]
\centering
\small 
\setlength{\tabcolsep}{5.pt} 
\renewcommand{\arraystretch}{1.2} 
\begin{tabular}{lllll}
\toprule
\multirow{2}{*}{$\mathcal{D'}$} & \multicolumn{2}{c}{C4}         & \multicolumn{2}{c}{Case-law}   
\\ 
\cmidrule(r){2-3} 
\cmidrule{4-5} 
                                & $\approx 1$Gb & $\approx 10$Gb & $\approx 1$Gb & $\approx 10$Gb \\
                                \midrule
Pythia-6.9B                     & 0.688         & \textbf{0.698} & 0.687         & 0.688          \\ \hline
\end{tabular}
\caption{AUC scores of DC-PDD in different reference corpus settings.}
\label{table4}
\vspace{-4mm}
\end{table}

\heading{Hyperparameter $a$} Recall that we set a hyperparameter $a$ to prevent the final score from being dominated by a few tokens, w.r.t. Eq.~\eqref{eq:7}.
We evaluate DC-PDD with different $a$ settings to investigate their impact on detection performance.
As shown in Table \ref{table5}, performance varies significantly with $a$ set to $0.001$, $0.01$, $0.1$, $1$, and $10$.
Actually, if $a$ is set too high, it does not effectively limit the calibrated token probabilities.
Conversely, if set too low, it will result in nearly equal calibrated token probabilities, causing scores for training and non-training text to be similar and thus, ineffective for detection.
From the Table \ref{table5}, we can see that the optimal $a$ setting varies across different target models and benchmarks.
For instance, the optimal $a$ is $10$ in detecting BookMIA from GPT-3 while it is $0.01$ in detecting PatentMIA from Qwen1.5-14B.
When $a$ is set to $0.01$, the overall performance for all models is optimal. Therefore, we recommend setting $a$ to $0.01$ when using DC-PDD for pretraining data detection in practical scenarios.
In future work, we will explore more flexible methods for setting $a$ to achieve better performance of DC-PDD.

\begin{table}[t]
\centering
\small 
\setlength{\tabcolsep}{5.pt} 
\renewcommand{\arraystretch}{1.2} 
\begin{tabular}{llllll}
\toprule
$a$                   & \multicolumn{1}{c}{0.001} & \multicolumn{1}{c}{0.01} & \multicolumn{1}{c}{0.1} & \multicolumn{1}{c}{1} & \multicolumn{1}{c}{10} \\ 
\midrule
\textbf{PatentMIA:} &                           &                          &                         &                       &                        \\
\multicolumn{1}{r}{Baichuan-13B}        & 0.645                     & \textbf{0.699}           & 0.664                   & 0.647                 & 0.645                  \\
\multicolumn{1}{r}{Qwen1.5-14B}          & 0.640                     & \textbf{0.689}           & 0.652                   & 0.623                 & 0.619                  \\ \hline
\textbf{BookMIA:}   &                           &                          &                         &                       &                        \\
GPT-3               & 0.673                     & 0.676                    & 0.665                   & 0.667                 & \textbf{0.725}          \\ 
\bottomrule       
\end{tabular}
\caption{AUC scores of DC-PDD in different $a$ settings.}
\label{table5}
\vspace{-4mm}
\end{table}

\vspace{-1mm}
\section{Related Work}

\vspace{-1mm}
\textbf{Membership inference attack (MIA).}
MIA is the de-facto threat model when evaluating privacy concerns in machine learning models.
First introduced by \citet{FirstMIA}, MIA's objective is to ascertain whether a specific sample was part of a model’s training dataset.
Prior MIA research has focused on traditional deep learning models \cite{MLMIA2, MLMIA3} and fine-tuning language models \cite{FTMIA1, FIMIA2, FIMIA3}.
But recently, MIA on LLMs has attracted growing attention with various applications, including examination of training data memorization \cite{MIA4M}, data contamination \cite{MIA4DC}, and copyright infringement \cite{MIA4CI1, MIA4CI2}. 
We consider a different type of MIA: pretraining data detection.

\heading{Pretraining data detection for LLMs}
Here, the MIA problem centers on identifying whether a piece of text was used by an LLM for pretraining. 
According to the access conditions to LLMs, current pretraining data detection methods for LLMs can be divided into two categories: (i) The white-box setting: assuming one has access to internals of LLMs, such as weights and activations. (ii) The black-box setting: assuming one can only query LLMs to compute token probabilities for the text.

There is limited research on the white-box setting since the internals of LLMs are typically not disclosed, rendering detection methods in white-box scenarios impractical.
\citet{wb} propose to use the probing technique for pretraining data detection, based on the assumption that texts encountered during the LLM's pretraining phase are represented differently in its internal activations compared to unseen texts.

Most research focuses on the black-box setting, assuming that the token probability distribution of a text can provide crucial information about whether the text was included in the training set. 
\citet{Extract} considered the model's perplexity for a text as an indicator to detect pretraining data from GPT-2 \cite{GPT2}.
They further introduced three methods, Zlib, Lowercase, and Smaller Ref, that take into account the intrinsic complexity of the target text.
More recently, \citet{Min-k} have proposed a straightforward yet well-performing method called Min-K\% Prob.
Min-K\% Prob tends to classify a non-training text composed of common words as training data.
A concurrent study Min-K\%++ Prob \cite{Min-k+} improves Min-K\% Prob by normalizing token probabilities, but requires access to the next-token prediction probability distribution across the LLM’s entire vocabulary, which is unavailable in closed-source LLMs like GPT-3 \cite{GPT3}.

We consider the black-box setting and calibrate the token probabilities before using them for detection.
What distinguishes our approach is that it neither requires additional reference models (unlike Small Ref) nor does it have extra access requirements on the LLM (unlike Min-K\%++ Prob). 

\vspace{-1mm}
\section{Conclusion}

\vspace{-1mm}
In this work, we proposed DC-PDD to improve methods that directly rely on token probabilities for pretraining data detection, which tend to misclassify non-training texts containing many common words as training texts.
The key idea of DC-PDD is to calibrate the token probabilities and thereby make them more informative signals for detection.
The calibration process is achieved by computing the cross-entropy (i.e., the divergence) between the token probability distribution and the token frequency distribution.
Experiments demonstrate the superior performances of DC-PDD compared to various baselines.
In future work, we want to detect whether an LLM was pretrained on a given corpus (corpus-level detection), rather than just on a piece of text (sample-level detection).


\section*{Limitations}

DC-PDD, while showing promising results in pre-training data detection from LLMs, has several limitations.
\begin{enumerate*}[label=(\roman*)]

    \item DC-PDD utilizes a reference corpus to calculate the token frequency distribution to estimate that of the training corpus. Although working, the similarity between these two distributions remains uncertain. Additionally, the language of reference corpus should be the same as that of text to be detected.

    \item Secondly, an important hyperparameter in DC-PDD is the upper bound of calibrated token probabilities. We have demonstrated its significant impact on method performance, but not how the optimal value should be set. We leave this issue to future work.

    \item Thirdly, DC-PDD is specific to textual data. While some detection methods can be applied universally across different data modalities by relying on sample-level loss values obtained from models, our method is based on token-level probability. This specificity hinders its direct application to other types of data, such as images.

    \item Fourthly, DC-PDD requires access to token probabilities, and therefore is not applicable to some closed-source models. In the future, we will explore detection methods based solely on model output to design more generalizable detection methods.
    
    \item Lastly, except for the closed-source model GPT-3 \cite{GPT3}, our research primarily focused on models with up to 20 billion parameters due to hardware constraints. Further studies replicating our work using larger-scale models will be essential to validate the effectiveness of DC-PDD in scenarios involving larger models.

\end{enumerate*}
\section*{Ethical Considerations}

Although DC-PDD aims to address issues such as copyright infringement or data contamination through pretraining data detection, it can also be used to compromise the privacy of individuals whose data has been used to train models, as pretraining data detection problem is an instance of Membership Inference Attacks (MIAs). Recognizing the potential risks associated with MIAs, we are extremely cautious with the data we use to ensure there is limited risk of any exposure of confidential data. For example, the PatentMIA benchmark is collected from the publicly available Google-Patents website and does not involve personal privacy data. Additionally, the other benchmarks we use have also been employed in prior research and do not pose any privacy risks.
\section*{Acknowledgements}

This work was funded by the National Natural Science Foundation of China (NSFC) under Grants No. 62472408 and 62372431, the Strategic Priority Research Program of the CAS under Grants No. XDB0680102 and XDB0680301, the National Key Research and Development Program of China under Grants No. 2023YFA1011602 and 2021QY1701, the Youth Innovation Promotion Association CAS under Grants No. 2021100, the Lenovo-CAS Joint Lab Youth Scientist Project, and the project under Grants No. JCKY2022130C039. 
This work was also (partially) funded by the Dutch Research Council (NWO), under project numbers 024.004.022, NWA.1389.20.\-183, and KICH3.LTP.20.006, and the European Union's Horizon Europe program under grant agreement No 101070212.All content represents the opinion of the authors, which is not necessarily shared or endorsed by their respective employers and/or sponsors.

\bibliography{references}

\begin{thebibliography}{41}
\expandafter\ifx\csname natexlab\endcsname\relax\def\natexlab#1{#1}\fi

\bibitem[{Achiam et~al.(2023)Achiam, Adler, Agarwal, Ahmad, Akkaya, Aleman, Almeida, Altenschmidt, Altman, Anadkat et~al.}]{GPT4}
Josh Achiam, Steven Adler, Sandhini Agarwal, Lama Ahmad, Ilge Akkaya, Florencia~Leoni Aleman, Diogo Almeida, Janko Altenschmidt, Sam Altman, Shyamal Anadkat, et~al. 2023.
\newblock {GPT-4} technical report.
\newblock \emph{arXiv preprint arXiv:2303.08774}.

\bibitem[{Amati and van Rijsbergen(2002)}]{DFR}
Gianni Amati and Cornelis~Joost van Rijsbergen. 2002.
\newblock Probabilistic models of information retrieval based on measuring the divergence from randomness.
\newblock \emph{ACM Transactions on Information Systems (TOIS)}, 20(4):357--389.

\bibitem[{Bai et~al.(2023)Bai, Bai, Chu, Cui, Dang, Deng, Fan, Ge, Han, Huang et~al.}]{Qwen}
Jinze Bai, Shuai Bai, Yunfei Chu, Zeyu Cui, Kai Dang, Xiaodong Deng, Yang Fan, Wenbin Ge, Yu~Han, Fei Huang, et~al. 2023.
\newblock Qwen technical report.
\newblock \emph{arXiv preprint arXiv:2309.16609}.

\bibitem[{Biderman et~al.(2023)Biderman, Schoelkopf, Anthony, Bradley, O’Brien, Hallahan, Khan, Purohit, Prashanth, Raff et~al.}]{Pythia}
Stella Biderman, Hailey Schoelkopf, Quentin~Gregory Anthony, Herbie Bradley, Kyle O’Brien, Eric Hallahan, Mohammad~Aflah Khan, Shivanshu Purohit, USVSN~Sai Prashanth, Edward Raff, et~al. 2023.
\newblock Pythia: A suite for analyzing large language models across training and scaling.
\newblock In \emph{International Conference on Machine Learning}, pages 2397--2430. PMLR.

\bibitem[{Black et~al.(2022)Black, Biderman, Hallahan, Anthony, Gao, Golding, He, Leahy, McDonell, Phang et~al.}]{GPT-neox}
Sid Black, Stella Biderman, Eric Hallahan, Quentin Anthony, Leo Gao, Laurence Golding, Horace He, Connor Leahy, Kyle McDonell, Jason Phang, et~al. 2022.
\newblock {GPT-NeoX-20B}: An open-source autoregressive language model.
\newblock \emph{arXiv preprint arXiv:2204.06745}.

\bibitem[{Brown et~al.(2020)Brown, Mann, Ryder, Subbiah, Kaplan, Dhariwal, Neelakantan, Shyam, Sastry, Askell et~al.}]{GPT3}
Tom Brown, Benjamin Mann, Nick Ryder, Melanie Subbiah, Jared~D Kaplan, Prafulla Dhariwal, Arvind Neelakantan, Pranav Shyam, Girish Sastry, Amanda Askell, et~al. 2020.
\newblock Language models are few-shot learners.
\newblock \emph{Advances in Neural Information Processing Systems}, 33:1877--1901.

\bibitem[{Cao et~al.(2024)Cao, Zhang, and Cheung}]{CDuse2}
Jialun Cao, Wuqi Zhang, and Shing-Chi Cheung. 2024.
\newblock Concerned with data contamination? {Assessing} countermeasures in code language model.
\newblock \emph{arXiv preprint arXiv:2403.16898}.

\bibitem[{Carlini et~al.(2022)Carlini, Chien, Nasr, Song, Terzis, and Tramer}]{LiRA}
Nicholas Carlini, Steve Chien, Milad Nasr, Shuang Song, Andreas Terzis, and Florian Tramer. 2022.
\newblock Membership inference attacks from first principles.
\newblock In \emph{2022 IEEE Symposium on Security and Privacy (SP)}, pages 1897--1914. IEEE.

\bibitem[{Carlini et~al.(2021)Carlini, Tramer, Wallace, Jagielski, Herbert-Voss, Lee, Roberts, Brown, Song, Erlingsson et~al.}]{Extract}
Nicholas Carlini, Florian Tramer, Eric Wallace, Matthew Jagielski, Ariel Herbert-Voss, Katherine Lee, Adam Roberts, Tom Brown, Dawn Song, Ulfar Erlingsson, et~al. 2021.
\newblock Extracting training data from large language models.
\newblock In \emph{30th USENIX Security Symposium (USENIX Security 21)}, pages 2633--2650.

\bibitem[{Chang et~al.(2023)Chang, Cramer, Soni, and Bamman}]{cpuse2}
Kent~K. Chang, Mackenzie Cramer, Sandeep Soni, and David Bamman. 2023.
\newblock Speak, memory: An archaeology of books known to {ChatGPT/GPT-4}.
\newblock \emph{arXiv preprint arXiv:2305.00118}.

\bibitem[{Chen et~al.(2023)Chen, Jian, Xi, Yi, Ding, Du, Zhu, Zong, Wang, and Zhang}]{CWT}
Jianghao Chen, Pu~Jian, Tengxiao Xi, Yidong Yi, Chenglin Ding, Qianlong Du, Guibo Zhu, Chengqing Zong, Jinqiao Wang, and Jiajun Zhang. 2023.
\newblock Chinesewebtext: Large-scale high-quality chinese web text extracted with effective evaluation model.
\newblock \emph{arXiv preprint arXiv:2311.01149}.

\bibitem[{Dong et~al.(2024)Dong, Jiang, Liu, Jin, and Li}]{CDuse1}
Yihong Dong, Xue Jiang, Huanyu Liu, Zhi Jin, and Ge~Li. 2024.
\newblock Generalization or memorization: Data contamination and trustworthy evaluation for large language models.
\newblock \emph{arXiv preprint arXiv:2402.15938}.

\bibitem[{Duan et~al.(2024)Duan, Suri, Mireshghallah, Min, Shi, Zettlemoyer, Tsvetkov, Choi, Evans, and Hajishirzi}]{LLMMIAA}
Michael Duan, Anshuman Suri, Niloofar Mireshghallah, Sewon Min, Weijia Shi, Luke Zettlemoyer, Yulia Tsvetkov, Yejin Choi, David Evans, and Hannaneh Hajishirzi. 2024.
\newblock Do membership inference attacks work on large language models?
\newblock \emph{arXiv preprint arXiv:2402.07841}.

\bibitem[{Duarte et~al.(2024)Duarte, Zhao, Oliveira, and Li}]{MIA4CI2}
Andr{\'e}~V Duarte, Xuandong Zhao, Arlindo~L Oliveira, and Lei Li. 2024.
\newblock De-cop: Detecting copyrighted content in language models training data.
\newblock \emph{arXiv preprint arXiv:2402.09910}.

\bibitem[{Google(2006)}]{GooglePatents}
Google. 2006.
\newblock Google {Patents}.
\newblock \href{https://patents.google.com/}{https://patents.google. com/}.

\bibitem[{Grynbaum and Mac(2023)}]{Openaisuit}
Michael~M. Grynbaum and Ryan Mac. 2023.
\newblock The {Times} sues {OpenAI} and {Microsoft} over {A.I}. use of copyrighted work.
\newblock \href{https://www.nytimes.com/2023/12/27/business/media/new-york-times-open-ai-microsoft-lawsuit.html}.

\bibitem[{Hisamoto et~al.(2020)Hisamoto, Post, and Duh}]{FTMIA1}
Sorami Hisamoto, Matt Post, and Kevin Duh. 2020.
\newblock Membership inference attacks on sequence-to-sequence models: Is my data in your machine translation system?
\newblock \emph{Transactions of the Association for Computational Linguistics}, 8:49--63.

\bibitem[{Jagannatha et~al.(2021)Jagannatha, Rawat, and Yu}]{FIMIA2}
Abhyuday Jagannatha, Bhanu Pratap~Singh Rawat, and Hong Yu. 2021.
\newblock Membership inference attack susceptibility of clinical language models.
\newblock \emph{arXiv preprint arXiv:2104.08305}.

\bibitem[{Jiang et~al.(2019)Jiang, Ren, Monz, and de~Rijke}]{CeLoss}
Shaojie Jiang, Pengjie Ren, Christof Monz, and Maarten de~Rijke. 2019.
\newblock Improving neural response diversity with frequency-aware cross-entropy loss.
\newblock In \emph{The World Wide Web Conference}, pages 2879--2885.

\bibitem[{Liu et~al.(2024)Liu, Zhu, Tan, Lu, Liu, and Chen}]{wb}
Zhenhua Liu, Tong Zhu, Chuanyuan Tan, Haonan Lu, Bing Liu, and Wenliang Chen. 2024.
\newblock Probing language models for pre-training data detection.
\newblock \emph{arXiv preprint arXiv:2406.01333}.

\bibitem[{Louis Brulé~Naudet(2024)}]{case-law}
Timothy~Dolan Louis Brulé~Naudet. 2024.
\newblock The case-law, centralizing legal decisions for better use.

\bibitem[{Mattern et~al.(2023)Mattern, Mireshghallah, Jin, Sch{\"o}lkopf, Sachan, and Berg-Kirkpatrick}]{FIMIA3}
Justus Mattern, Fatemehsadat Mireshghallah, Zhijing Jin, Bernhard Sch{\"o}lkopf, Mrinmaya Sachan, and Taylor Berg-Kirkpatrick. 2023.
\newblock Membership inference attacks against language models via neighbourhood comparison.
\newblock \emph{arXiv preprint arXiv:2305.18462}.

\bibitem[{Meeus et~al.(2023)Meeus, Jain, Rei, and de~Montjoye}]{MIA4CI1}
Matthieu Meeus, Shubham Jain, Marek Rei, and Yves-Alexandre de~Montjoye. 2023.
\newblock Did the neurons read your book? {Document-level} membership inference for large language models.
\newblock \emph{arXiv preprint arXiv:2310.15007}.

\bibitem[{Mozes et~al.(2023)Mozes, He, Kleinberg, and Griffin}]{Priuse}
Maximilian Mozes, Xuanli He, Bennett Kleinberg, and Lewis~D. Griffin. 2023.
\newblock Use of {LLMs} for illicit purposes: Threats, prevention measures, and vulnerabilities.
\newblock \emph{arXiv preprint arXiv:2308.12833}.

\bibitem[{Nasr et~al.(2023)Nasr, Carlini, Hayase, Jagielski, Cooper, Ippolito, Choquette-Choo, Wallace, Tram{\`e}r, and Lee}]{MIA4M}
Milad Nasr, Nicholas Carlini, Jonathan Hayase, Matthew Jagielski, A~Feder Cooper, Daphne Ippolito, Christopher~A Choquette-Choo, Eric Wallace, Florian Tram{\`e}r, and Katherine Lee. 2023.
\newblock Scalable extraction of training data from (production) language models.
\newblock \emph{arXiv preprint arXiv:2311.17035}.

\bibitem[{Oren et~al.(2023)Oren, Meister, Chatterji, Ladhak, and Hashimoto}]{MIA4DC}
Yonatan Oren, Nicole Meister, Niladri Chatterji, Faisal Ladhak, and Tatsunori~B Hashimoto. 2023.
\newblock Proving test set contamination in black box language models.
\newblock \emph{arXiv preprint arXiv:2310.17623}.

\bibitem[{Radford et~al.(2019)Radford, Wu, Child, Luan, Amodei, Sutskever et~al.}]{GPT2}
Alec Radford, Jeffrey Wu, Rewon Child, David Luan, Dario Amodei, Ilya Sutskever, et~al. 2019.
\newblock Language models are unsupervised multitask learners.
\newblock \emph{OpenAI blog}, 1(8):9.

\bibitem[{Raffel et~al.(2020)Raffel, Shazeer, Roberts, Lee, Narang, Matena, Zhou, Li, and Liu}]{C4}
Colin Raffel, Noam Shazeer, Adam Roberts, Katherine Lee, Sharan Narang, Michael Matena, Yanqi Zhou, Wei Li, and Peter~J Liu. 2020.
\newblock Exploring the limits of transfer learning with a unified text-to-text transformer.
\newblock \emph{Journal of Machine Learning Research}, 21(140):1--67.

\bibitem[{Sablayrolles et~al.(2019)Sablayrolles, Douze, Schmid, Ollivier, and J{\'e}gou}]{MLMIA2}
Alexandre Sablayrolles, Matthijs Douze, Cordelia Schmid, Yann Ollivier, and Herv{\'e} J{\'e}gou. 2019.
\newblock White-box vs black-box: Bayes optimal strategies for membership inference.
\newblock In \emph{International Conference on Machine Learning}, pages 5558--5567. PMLR.

\bibitem[{Shi et~al.(2024)Shi, Ajith, Xia, Huang, Liu, Blevins, Chen, and Zettlemoyer}]{Min-k}
Weijia Shi, Anirudh Ajith, Mengzhou Xia, Yangsibo Huang, Daogao Liu, Terra Blevins, Danqi Chen, and Luke Zettlemoyer. 2024.
\newblock \href {https://openreview.net/forum?id=zWqr3MQuNs} {Detecting pretraining data from large language models}.
\newblock In \emph{The Twelfth International Conference on Learning Representations}.

\bibitem[{Shokri et~al.(2017)Shokri, Stronati, Song, and Shmatikov}]{FirstMIA}
Reza Shokri, Marco Stronati, Congzheng Song, and Vitaly Shmatikov. 2017.
\newblock Membership inference attacks against machine learning models.
\newblock In \emph{2017 IEEE symposium on security and privacy (SP)}, pages 3--18. IEEE.

\bibitem[{Song and Shmatikov(2019)}]{MLMIA3}
Congzheng Song and Vitaly Shmatikov. 2019.
\newblock Auditing data provenance in text-generation models.
\newblock In \emph{Proceedings of the 25th ACM SIGKDD International Conference on Knowledge Discovery \& Data Mining}, pages 196--206.

\bibitem[{Stempel(2024)}]{Nvidiasuit}
Jonathan Stempel. 2024.
\newblock Nvidia is sued by authors over {AI} use of copyrighted works.
\newblock \href{https://www.reuters.com/technology/nvidia-is-sued-by-authors-over-ai-use-copyrighted-works-2024-03-10/}.

\bibitem[{Team(2024)}]{Qwen1.5}
Qwen Team. 2024.
\newblock \href {https://qwenlm.github.io/blog/qwen1.5/} {Introducing qwen1.5}.

\bibitem[{Touvron et~al.(2023{\natexlab{a}})Touvron, Lavril, Izacard, Martinet, Lachaux, Lacroix, Rozi{\`e}re, Goyal, Hambro, Azhar et~al.}]{Llama}
Hugo Touvron, Thibaut Lavril, Gautier Izacard, Xavier Martinet, Marie-Anne Lachaux, Timoth{\'e}e Lacroix, Baptiste Rozi{\`e}re, Naman Goyal, Eric Hambro, Faisal Azhar, et~al. 2023{\natexlab{a}}.
\newblock Llama: Open and efficient foundation language models.
\newblock \emph{arXiv preprint arXiv:2302.13971}.

\bibitem[{Touvron et~al.(2023{\natexlab{b}})Touvron, Martin, Stone, Albert, Almahairi, Babaei, Bashlykov, Batra, Bhargava, Bhosale et~al.}]{Llama2}
Hugo Touvron, Louis Martin, Kevin Stone, Peter Albert, Amjad Almahairi, Yasmine Babaei, Nikolay Bashlykov, Soumya Batra, Prajjwal Bhargava, Shruti Bhosale, et~al. 2023{\natexlab{b}}.
\newblock Llama 2: Open foundation and fine-tuned chat models.
\newblock \emph{arXiv preprint arXiv:2307.09288}.

\bibitem[{Watson et~al.(2021)Watson, Guo, Cormode, and Sablayrolles}]{RefMIA}
Lauren Watson, Chuan Guo, Graham Cormode, and Alex Sablayrolles. 2021.
\newblock On the importance of difficulty calibration in membership inference attacks.
\newblock \emph{arXiv preprint arXiv:2111.08440}.

\bibitem[{Yang et~al.(2023)Yang, Xiao, Wang, Zhang, Bian, Yin, Lv, Pan, Wang, Yan et~al.}]{Baichuan}
Aiyuan Yang, Bin Xiao, Bingning Wang, Borong Zhang, Ce~Bian, Chao Yin, Chenxu Lv, Da~Pan, Dian Wang, Dong Yan, et~al. 2023.
\newblock Baichuan 2: Open large-scale language models.
\newblock \emph{arXiv preprint arXiv:2309.10305}.

\bibitem[{Yeom et~al.(2018)Yeom, Giacomelli, Fredrikson, and Jha}]{MLMIA1}
Samuel Yeom, Irene Giacomelli, Matt Fredrikson, and Somesh Jha. 2018.
\newblock Privacy risk in machine learning: Analyzing the connection to overfitting.
\newblock In \emph{2018 IEEE 31st Computer Security Foundations Symposium (CSF)}, pages 268--282. IEEE.

\bibitem[{Zhang et~al.(2024)Zhang, Sun, Yeats, Ouyang, Kuo, Zhang, Yang, and Li}]{Min-k+}
Jingyang Zhang, Jingwei Sun, Eric Yeats, Yang Ouyang, Martin Kuo, Jianyi Zhang, Hao Yang, and Hai Li. 2024.
\newblock {Min-K\%++}: Improved baseline for detecting pre-training data from large language models.
\newblock \emph{arXiv preprint arXiv:2404.02936}.

\bibitem[{Zhang et~al.(2022)Zhang, Roller, Goyal, Artetxe, Chen, Chen, Dewan, Diab, Li, Lin et~al.}]{OPT}
Susan Zhang, Stephen Roller, Naman Goyal, Mikel Artetxe, Moya Chen, Shuohui Chen, Christopher Dewan, Mona Diab, Xian Li, Xi~Victoria Lin, et~al. 2022.
\newblock {OPT}: Open pre-trained transformer language models.
\newblock \emph{arXiv preprint arXiv:2205.01068}.

\end{thebibliography}
\appendix
\section{Appendix}

\subsection{Baseline details}\label{appendix-baselines}
The baselines are all based on a detection score to determine a text $x$ whether was included in the per-training corpus of an LLM $\mathcal{M}$. Followings are the details of how they calculate the detection score.

\heading{PPL} \cite{Extract} This is an instance of Loss Attack proposed by \cite{MLMIA1}. In the context of LLMs, this loss corresponds to perplexity. Thus, the detection score is the perplexity of $x$.  A low score suggests that $x$ was likely part of the pretraining data.

\heading{Small Ref} \cite{Extract} This method exactly follows the approach described by \citet{RefMIA}, which assumes access to a reference model, $\mathcal{M}_{ref}$, trained on a disjoint set of training data drawn from a similar distribution and posits that the intrinsic complexity of $x$ can be quantified as $\mathcal{M}_{ref}$'s perplexity for $x$. Since the assumption is impractical, the Small Ref method employs a smaller model from the same family of $\mathcal{M}$ as a substitute for $\mathcal{M}_{ref}$, and then calibrate $\mathcal{M}$'s perplexity for $x$ using a difficulty estimate through the smaller model's perplexity for $x$. Consequently, the detection score is calculated as the ratio of $x$'s perplexity under $\mathcal{M}$ to $x$'s perplexity under a smaller model pre-trained on the same data. A low score suggests that $x$ was likely part of the pretraining data.

\heading{Zlib} \cite{Extract} Similar to the Small Ref method, but uses the zlib entropy of $x$ in place of the smaller model’s perplexity for $x$. The zlib entropy is the entropy in bits when the sequence is compressed using \textit{zlib}.\footnote{\url{https://github.com/madler/zlib}} The detection score is then determined by the ratio of $\mathcal{M}$’s perplexity for $x$ to the zlib entropy of $x$. A low score suggests that $x$ was likely part of the pretraining data.

\heading{Lowercase} \cite{Extract} Similarly to the Small Ref method, but uses $\mathcal{M}$'s perplexity for the lowercase of $x$ to replace the smaller model’s perplexity for $x$. The detection score is then determined by the ratio of $\mathcal{M}$’s perplexity for $x$ to $\mathcal{M}$’s perplexity for the lowercase of $x$. A low score suggests that $x$ was likely part of the pretraining data.

\heading{Min-K\% Prob} \cite{Min-k} Min-K\% Prob is based on the intuition that non-member examples tend to have more tokens assigned lower probabilities than member examples do. Thus, it begins by calculating the probability of each token in $x$, then selects the k\% of tokens with the lowest probabilities to compute their average log-likelihood as the detection score. A high score suggests that $x$ was likely part of the pretraining data.

\heading{Min-K\%++ Prob} \cite{Min-k+} The underlying idea of Min-K\%++ Prob is that if the probability of the current input token surpasses the probabilities of other tokens in the vocabulary, it is probable that the input has been seen during training, irrespective of the actual probability value of the input token. Therefore, it first calculates the probability of each token in $x$, then normalizes the token probability using the statistics of the categorical distribution over the entire vocabulary, and finally selects the $k$\% of tokens with the lowest normalized probabilities to compute their average as the detection score. A high score suggests that $x$ was likely part of the pretraining data.

\subsection{Metrics}\label{appendix-metrics}
\heading{Area Under the ROC Curve (AUC)} The AUC score quantifies the overall performance of a classification method. To calculate the AUC score for a method, we need to compute the True Positive Rates (TPRs) and False Positive Rates (FPRs) at all classification thresholds and plot a TPR vs. FPR curve, known as the ROC curve. The AUC is then defined as the Area Under the ROC curve, providing an aggregate measure of the effect of all possible classification thresholds. Therefore, AUC provides a comprehensive, threshold-independent score that reflects the method's ability to distinguish between positive and negative cases effectively.

\heading{TPR (true positive rate) at a low FPR (false positive rate)} We report TPR at a low FPR by adjusting the threshold value, Specifically, we choose $5$\% as our target FPR value, and report the corresponding TPR value.

\end{document}